\documentclass{article}
\usepackage{amssymb}

\usepackage[preprint]{corl_2025} 

\usepackage{hyperref}
\usepackage{url}

\usepackage{multirow}
\usepackage{booktabs}
\usepackage{stfloats} 
\usepackage{graphicx}
\usepackage{xcolor}
\usepackage[export]{adjustbox}

\usepackage{amsmath}
\usepackage{amssymb}
\usepackage{mathtools}

\usepackage{algorithm}
\usepackage{algpseudocode}
\usepackage{color}
\usepackage{wrapfig}

\newcommand{\bx}{{\bf x}}
\newcommand{\bX}{{\bf X}}

\newcommand{\bs}{{\bf s}}

\newcommand{\cD}{\mathcal{D}}

\newcommand{\cL}{{\mathcal L}}

\newcommand{\fref}[1] {Fig.~\ref{#1}}

\title{Improving Efficiency of Sampling-based Motion Planning via Message-Passing Monte Carlo}

\author{
    Makram Chahine \\
  CSAIL, MIT \\
  \texttt{chahine@mit.edu} \\
  \And
  T. Konstantin Rusch\\
  ELLIS Institute T\"{u}bingen \&\\
  Max Planck Institute for Intelligent Systems \&\\
  T\"{u}bingen AI Center\\
  \texttt{tkrusch@tue.ellis.eu} \\
  \AND
  Zach J. Patterson \\
  Case Western Reserve University\\ Mechanical and Aerospace Engineering \\
  \texttt{zpatt@case.edu} \\
  \And
  Daniela Rus \\
  CSAIL, MIT \\
  \texttt{rus@csail.mit.edu} \\
}

\begin{document}
\maketitle


\begin{abstract}
Sampling-based motion planning methods, while effective in high-dimensional spaces, often suffer from inefficiencies due to irregular sampling distributions, leading to suboptimal exploration of the configuration space. In this paper, we propose an approach that enhances the efficiency of these methods by utilizing low-discrepancy distributions generated through Message-Passing Monte Carlo (MPMC). MPMC leverages Graph Neural Networks (GNNs) to generate point sets that uniformly cover the space, with uniformity assessed using the the $\cL_p$-discrepancy measure, which quantifies the irregularity of sample distributions. By improving the uniformity of the point sets, our approach significantly reduces computational overhead and the number of samples required for solving motion planning problems. Experimental results demonstrate that our method outperforms traditional sampling techniques in terms of planning efficiency.
\end{abstract}



\section{Introduction}
Sampling-based motion planning is a key method for robot navigation in complex environments. From Probabilistic Roadmaps (PRMs)~\citep{prm} to Rapidly-exploring Random Trees (RRTs)~\citep{rrt}, the focus of much of the research in this area has been on developing more efficient and optimal motion planning algorithms~\citep{janson2015fastmarchingtreefast, star} that can handle high-dimensional spaces~\citep{hidim, orthey2019rapidlyexploringquotientspacetreesmotion}, various types of constraints~\citep{Karaman2013SamplingbasedOM, constrain}, and dynamic environments~\citep{Phillips2011SIPPSI, Otte2016RRTXAO}. 

Instead of refining motion planning algorithms, our work targets the fundamental building block underlying these approaches—the core sampling process itself. We introduce a novel sampling strategy based on Message-Passing Monte Carlo (MPMC)~\citep{mpmc}, a graph neural network architecture trained to generate low-discrepancy point sets on unit hypercubes of arbitrary dimension. We expand the generic MPMC algorithm with the introduction of a novel training objective tailored to high-dimensional spaces, ensuring that generated points are optimally distributed and scalable. 

This leads to a potent unbiased state sampling technique, that can be seamlessly integrated into any sampling-based planner, and that requires neither conditioning on the workspace description, nor a steer function, nor past examples, nor start and goal information.
This approach provides strong theoretical guarantees and it also outperforms traditional techniques across various benchmarks, in environments of varying complexity and dimensionality.
\clearpage
Our contributions are stated as follows:
\begin{itemize}
    \item We introduce the first application of MPMC neural network point set generation in motion planning, a significant novelty for unbiased sampling techniques.
    \item We propose a novel training objective tailored to high-dimensional configuration spaces, ensuring the generated points are well-suited for complex planning problems.
    \item We support our approach with rigorous theoretical justification, linking the training objective of MPMC point sets to a tighter upper bound on the distance from the optimal path.
    \item We establish superior planning efficiency against the current gold standard sampling approach in motion planning on a variety of PRM benchmarks, including challenging high-dimensional environments.
    \item We demonstrate the effectiveness of our sampling technique on a real-world UR5 robot arm, showing its potential for practical deployment in robotics.
\end{itemize}

Our MPMC-based sampling strategy offers a powerful and versatile alternative to traditional sampling methods. By targeting the sampling process directly, we open up new possibilities for improving the efficiency of a wide array of motion planning algorithms in robotic systems.

\section{Related work}

\subsection{Learning for sampling-based planners}

The incorporation of machine learning into motion planning has opened up new ways to accelerate pathfinding by learning from past experiences. One common strategy involves storing and reusing previously computed paths or solutions. For example, methods such as path libraries~\citep{experience}, sparse roadmaps~\citep{coleman2014experiencebasedplanningsparseroadmap}, and local obstacle roadmaps~\citep{simobs} allow a robot to retrieve and adapt previously successful solutions to new, but similar, planning problems. These approaches reduce the computation time by narrowing the search space using knowledge from past instances.

Another line of research enhances the sampling process by learning distributions that guide planners toward more promising regions of the configuration space. Some methods employ problem-invariant distributions~\citep{repetition}, while others adapt based on the workspace environment~\citep{Chamzas_2021}. Deep learning further extends these concepts by learning from prior planning tasks, enabling distributions that condition on both the workspace and specific start-goal configurations of new problems~\citep{chamzas2022learningretrieverelevantexperiences}. These learned distributions effectively bias the sampling process, improving convergence rates and solution quality in new planning scenarios.

In contrast, our method departs from these past learning-based approaches by introducing a deterministic sampling strategy obtained via neural network training that does not depend on conditioning or prior knowledge.

\subsection{Low-discrepancy constructions}
Over the past century, numerous constructions of low-discrepancy point sets and sequences have been proposed. Most constructions are deeply rooted in number theory and abstract algebra. A widely used building block of low-discrepancy constructions is the one-dimensional van der Corput sequence~\citep{VDC1935} in base $b$, which is generalized to a higher dimensional setting via the Halton sequence ~\citep{HALTON1960}. Each of the $d$ coordinates in a Halton sequence corresponds to a distinct van der Corput sequence in base $b$, with the bases selected to be co-prime. Faure sequences, as introduced in~\citep{FAURE1982}, offer a similar construction to Halton sequences but incorporate permutations of the digits in base $b$.

Another broad class of low-discrepancy constructions are today known as digital $(t,s)$-sequences, first introduced in~\citep{NIED1987}. These constructions include the widely known Sobol sequence~\citep{SOBOL1967} which is constructed using tools from linear algebra involving primitive polynomials and well-chosen generating matrices defined over finite fields. 

A further distinct approach, rooted in a different branch of number theory, emerged with Korobov's introduction of the good-lattice method~\citep{KOROBOV1963}. This technique utilizes modular arithmetic and prime number properties to construct a structured, grid-like set of integration nodes. Since its introduction, lattice rules have undergone significant extensions and refinements. Comprehensive discussions and valuable reviews of these developments are available in~\citep{HABER1970, SLOANLATTICE1985, NUYENS2014, LATTICES2022}.

Several approaches have recently emerged targeting the construction of points for fixed dimension and number of points. In~\citep{DOERR2013}, new low-discrepancy point sets were proposed by optimizing permutations applied to the Halton sequence. Another approach, known as subset selection, was introduced to select $k < N$ points from an $N$-element set that minimize the discrepancy. An exact algorithm for this selection was presented in~\citep{CLEMENTDOERR2022}, while a swap-based heuristic approach was employed in~\citep{CLEMENTDOERR2024}. Additionally,~\citep{CLEMENTOPTIMAL2023} proposed a non-linear programming method to generate point sets with optimal star-discrepancy for fixed dimension and number of points. However, this approach faces significant computational challenges, limiting its practical application to finding optimal sets for only up to $21$ points in two dimensions and $8$ points in three dimensions. 
MPMC significantly differs from all previous approches by its explicit use of machine learning. Moreover, it has been shown in~\citep{mpmc} that MPMC generates point sets with significantly better distributional properties compared to any previous method, reaching optimal or near-optimal discrepancy.

\section{Methods}
\label{sec:methods}

Message-Passing Monte Carlo (MPMC) is a machine learning approach designed to generate low-discrepancy point sets, which are the key for efficiently covering space in a uniform manner. MPMC leverages Graph Neural Networks (GNNs) and tools from Geometric Deep Learning to generate these point sets. The method focuses on the geometric properties needed to ensure uniformity and it is highly versatile for generating points across different dimensions. In this section we describe MPMC and its extensions for the motion planning application domain. 

\subsection{Message-Passing Monte Carlo Sampling}
\begin{figure*}[t]

\includegraphics[width=0.9\textwidth]
{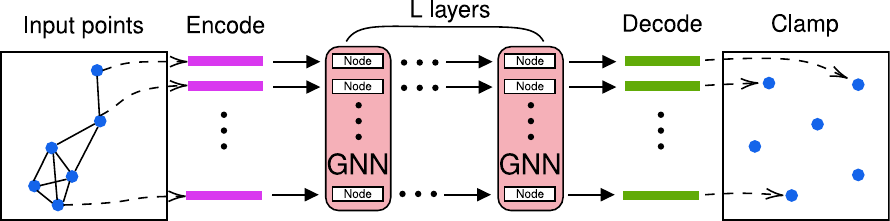}
\centering
\caption{Schematic of the MPMC model reproduced from~\citep{mpmc}. First, random input points are encoded to a high dimensional representation. Second, the encoded representations are passed through a deep GNN, where the underlying computational graph is constructed based on nearest neighbors using the positions of the initial input points. Finally, the node-wise output representations of the final GNN layer are decoded and clamped back into the unit hypercube.}
\label{fig:MPMC_model}
\end{figure*}

Message-Passing Monte Carlo (MPMC)~\citep{mpmc} leverages Graph Neural Networks (GNNs) to generate point sets that cover the space in a uniform manner (for an extension to non-uniform distributions see \citep{kirk2025low}). This uniformity can be assessed through measures of irregularity termed discrepancy. While there exist a plethora of different uniformity measures, we focus on the $\cL_p$-discrepancy here. That is, given a set of points $\{\bX_i\}_{i=1}^N$ in the unit hypercube $[0,1]^d$ and $p\geq1$, the $\cL_p$-discrepancy is defined as,
\begin{equation}
\begin{aligned}
\label{eq:L2_disc}
    \cL_p^p(\{\bX_i\}_{i=1}^N) := \int_{[0,1]^d} \left| \frac{\#(\{\bX_i\}_{i=1}^N \cap [0,\bx))}{N} - \mu([0,\bx)) \right|^p d\bx,
\end{aligned}
\end{equation}
where $\#(\{\bX_i\}_{i=1}^N \cap [0,\bx))$ counts how many points of $\{\bX_i\}_{i=1}^N$ fall inside the box $[0,\bx) = \prod_{i=1}^d [0,x_i)$ for $\bx = (x_1,\ldots,x_d)\in [0,1]^d$, and $\mu(\cdot)$ denotes the usual Lebesgue measure. A differentiable closed-form solution to the high-dimensional integral in \eqref{eq:L2_disc} for the case of $p=2$, known as Warnock's formula~\citep{warnock1972computational}, can be computed in $\mathcal{O}(N^2d)$. Consequently, MPMC utilizes the $\cL_2$-discrepancy as the training loss to generate point sets with low-discrepancy. The MPMC model transforming random input points into low-discrepancy sets via deep GNNs is depicted in \fref{fig:MPMC_model}.

\subsubsection{Fast MPMC training in higher dimensions}
As described in~\citep{mpmc}, the $\cL_2$-discrepancy fails to distinguish random from highly uniform points in high dimensions for moderate amount of points. Therefore, it was suggested in~\citep{mpmc} to minimize the Hickernell $\cL_2$-discrepancy $D_{H,2}$~\citep{hickernell} in this case, that sums over all $\cL_2$-discrepancies of $k$-dimensional projections, with $1\leq k \leq d$. A straightforward implementation of $D_{H,2}$ has a computational complexity of $\mathcal{O}((2^d-1)N^2d)$, which would involve evaluating the $\mathcal{L}_2$-discrepancy for billions of projections for $d\geq 30$. To mitigate this issue, the authors in~\citep{mpmc} limit the summation to smaller, randomly selected subsets of projections. In contrast, in this work we leverage the closed-form solution of the Hickernell $\cL_2$-discrepancy from~\citep{joe1997formulas} that can be computed again in $\mathcal{O}(N^2d)$ instead of $\mathcal{O}((2^d-1)N^2d)$ \citep{clement2025optimization}. More concretely, 
\begin{align*}
    D_{H,2}^2(\{\bX_i\}_{i=1}^N) &= \sum_{\emptyset\neq s \subseteq \{1,\dots,d\}} \cL^2_2(\{\bX^{(s)}_i\}_{i=1}^N) \\&= \left(\frac{4}{3}\right)^d - \frac{2}{N}\sum_{i=1}^N\prod_{k=1}^d\left(\frac{3}{2}-\frac{\bX_{i,k}^2}{2}\right) + \frac{1}{N^2}\sum_{i=1}^N \sum_{j=1}^N \prod_{k=1}^d [2-\max(\bX_{i,k},\bX_{j,k})],
\end{align*}
where $\bX_{i,k}$ is the $k$-th entry of $\bX_i$, and  $\{\bX^{(s)}_i\}_{i=1}^N$ is the projection of $\{\bX_i\}_{i=1}^N$ onto $[0,1]^{|s|}$. 
This enables very fast training in higher dimensions, minimizing the discrepancy over all projections. 

\subsection{Motion planning via Probabilistic Road Maps}

\begin{wrapfigure}{r}{0.5\textwidth}
    \vspace{-22pt} 
    \begin{minipage}{0.48\textwidth}
        \begin{algorithm}[H]
        \caption{ps-PRM Algorithm}
        \begin{algorithmic}[1]
        \State $S \gets \texttt{Sample}(N)$
        \State $S_v \gets \texttt{PruneInvalid}(S)$
        \State $V \gets \{\bX_{\text{start}}, \bX_{\text{goal}}\} \cup S_v$
        \State $E \gets \emptyset$
        \ForAll{$v \in V$}
            \State $P_{\text{near}} \gets \texttt{Near}(V \setminus \{v\}, v, r_N)$
            \ForAll{$p \in P_{\text{near}}$}
                \If{$\texttt{CollisionFree}(v, p)$}
                    \State $E \gets E \cup \{(v, p)\} \cup \{(p, v)\}$
                \EndIf
            \EndFor
        \EndFor
        \State \Return $\texttt{ShortestPath}(\bX_{\text{start}}, \bX_{\text{goal}}, V, E)$
        \end{algorithmic}
        \label{alg:psprm}
        \end{algorithm}
    \end{minipage}
    \vspace{-1pt} 
\end{wrapfigure}

In this work, the focus is on the efficiency of sampling-based motion planning in the sense of task success with respect to number of points sampled. Furthermore, for convenience in dealing with pre-trained sets of points of fixed number, we use a particular instantiation of the Probabilistic Roadmap (PRM) algorithm, which we refer to as \texttt{ps-PRM} (for pre-sampled PRM). This specific version of the PRM algorithm is described in Algorithm \ref{alg:psprm}. 

The algorithm begins by sampling $N$ points from the entire space to form the set $S$.
Next, we prune out milestones from the initial set $S$ that do not fall in free space, keeping only valid samples $S_v$, which are combined with the start and goal $\{\bX_{\text{start}}, \bX_{\text{goal}}\}$ to obtain $V$.
For each node $v$ in $V$, the algorithm identifies nearby nodes within a radius $r_N$, with the set of such points denoted $P_{\text{near}}$.
For each neighbor $p$ in $P_{\text{near}}$, the algorithm checks for collision-free paths between $v$ and $p$. If the path is free of collisions, a bidirectional edge is added between $v$ and $p$ to the edge set $E$.
Finally, the algorithm attempts to find a shortest path from $\bX_{\text{start}}$ to $\bX_{\text{goal}}$ 
and returns the path if one exists; otherwise, it indicates failure.

\subsection{Theoretical guarantee}
Next, we outline a theoretical justification of minimizing discrepancy as a means to improving efficiency of sampling-based motion planning. To this end, we introduce another uniformity measure known as dispersion, which is commonly used in assessing the efficiency of sampling-based motion planning. Given a point set $\{\bX^s_i\}_{i=1}^N$ in $[0,1]^d$ and $p\geq1$, the $l_p$-dispersion is then defined as,
\begin{align*}
    \cD_p(\{\bX_i\}_{i=1}^N)= \sup_{\bs \in [0,1]^d} \min_{1\leq i \leq N}\|\bs-\bX_i\|_p.
\end{align*}
Dispersion is closely related to discrepancy through the following inequality established in~\citep{niederreiter1992random},
\begin{align*}
\cD_\infty(\{\bX^s_i\}_{i=1}^N) \leq \cL_\infty(\{\bX^s_i\}_{i=1}^N)^{1/d},
\end{align*}
with the $\cL_\infty$-discrepancy. Based on this and following~\citep{dispersion_sampling}, one can provide a deterministic sampling guarantee. Namely, using the PRM planning algorithm, assuming some mild assumptions are satisfied (e.g., feasible path $\delta$-clearance), and choosing a radius $r_N$ of the PRM algorithm based on $N$ samples according to,
\begin{align*}
    r_N = 2\alpha \sqrt{d}\cL_\infty(\{\bX^s_i\}_{i=1}^N)^{\frac{1}{d}},
\end{align*}
for some $\alpha>1$ and $r_N$ satisfying conditions corresponding to the path $\delta$-clearance, one can guarantee that the cost of the returned path is within a factor of $\frac{1}{\alpha-1}$ of the optimal $\delta$-clear path. Clearly, a low-discrepancy induces a guarantee of the resulting path to be close to the optimal $\delta$-clear path.

\paragraph{Note.} The MPMC points are generated using the Hickernell $\cL_2$-discrepancy loss, while our guarantee is related to the $\cL_\infty$-discrepancy. Though the relation between $\cL_\infty$ and $\cL_2$ remains an open question in discrepancy theory, empirical evidence suggests the two are very close to each other at least for medium numbers of points \citep{mpmc}. Elsewhere, a clarification on $\cD_\infty$ dispersion used in the referenced result from \citep{dispersion_sampling} and its relationship to grids of points is provided in Appendix~\ref{app:disdis}.

\section{Experimental Evaluation}

We evaluate our motion planning solution in simulation, using a variety of environments ranging from 2D mazes to higher dimensional spaces and in a physical setting with a UR5 robot arm. We assess performance according to the number of sampled points and success rate in finding a path. 

\subsection{Sampling methods}

\begin{wrapfigure}{r}{0.5\textwidth}
    \vspace{-15pt}
    \begin{minipage}{0.5\textwidth}
            \includegraphics[width=\textwidth]{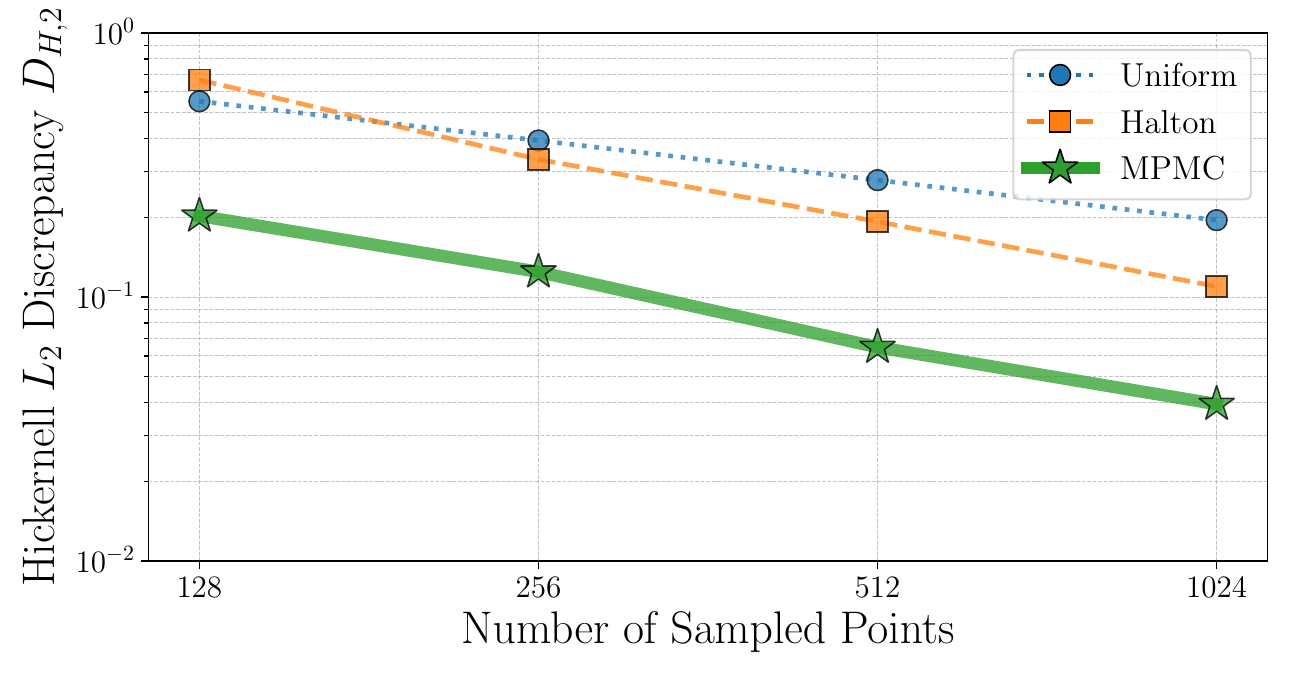}
            \vspace{-15pt}
            \caption{Hickernell $\cL_2$-discrepancy in $d=10$ for Uniform, Halton and MPMC sampling.}
            \vspace{-15pt}
            \label{fig:discreps}
    \end{minipage}
\end{wrapfigure}

The performance of our method is evaluated against Uniform sampling and also against Halton~\citep{halton} and Sobol~\citep{sobol} (for the 2-D maze setup only) sequences, two widely-used low-discrepancy Quasi-Monte Carlo methods (see Appendix~\ref{app:2dpts} for 2D-visualization). Sobol sequences are designed to efficiently cover spaces by minimizing gaps between sample points, making them ideal for complex problems. Halton sequences generate points using prime number bases, providing effective uniform coverage. However, both approaches suffer from correlation issues in higher dimensions~\citep{mpmc}. 

To obtain a statistical measure of performance across repeated runs, we randomize the selection of the MPMC points from the pool of trained point sets (batches can contain 8, 16 or 32 point sets based on the training instance). Similarly for the deterministic Halton and Sobol sequences, we initialize a new sequence where the previous one ended. Hence, when sampling point sets of size \( N \), we take the points indexed from \( (i-1) \ N + 1 \) to \( i*N \) for the \( i \)-th run, with \(i\) taking integer values from 1 to the number of runs fixed for the experiment.

To illustrate differences between the sampling techniques considered in the context of this work, we provide the Hickernell $\cL_2$-discrepancies of point sets of sizes $\{128, 256, 512, 1024\}$ in 10 dimensions in Figure \ref{fig:discreps}. Halton sequences never surpass a 2 times advantage over Uniform sampling, and can even offer worse discrepancy for smaller $N$. MPMC points, on the other hand, reach substantially lower Hickernell $\cL_2$-discrepancy, consistently close to 3 times lower than that of Halton and achieving up to a 5-fold improvement over Uniform sampling for larger point sets.

\subsection{Experiments}

\begin{figure}[t]
\centering
\includegraphics[width=0.3\textwidth,cfbox=gray 1pt 1pt]{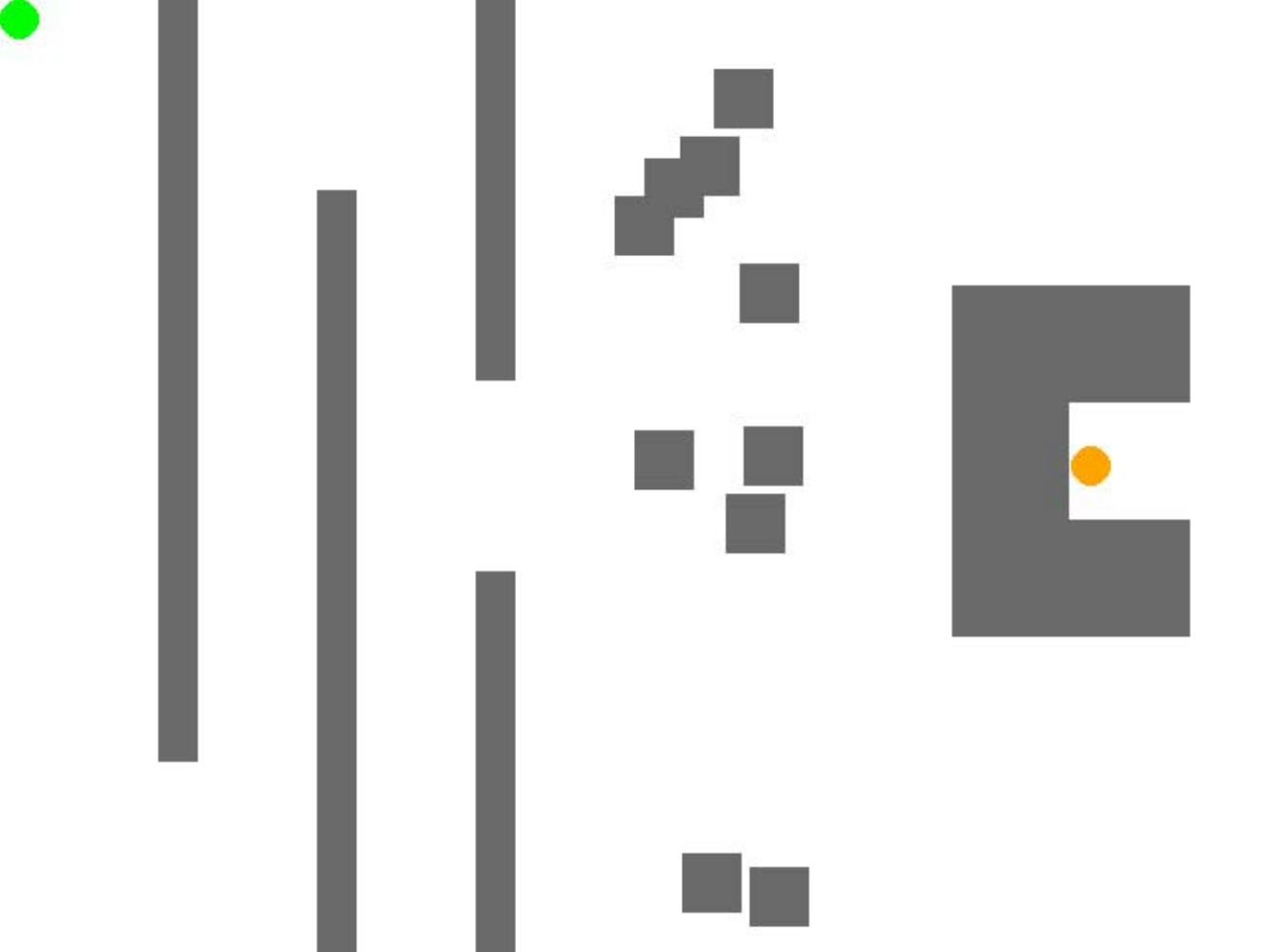}
\includegraphics[width=0.3\textwidth,cfbox=gray 1pt 1pt]{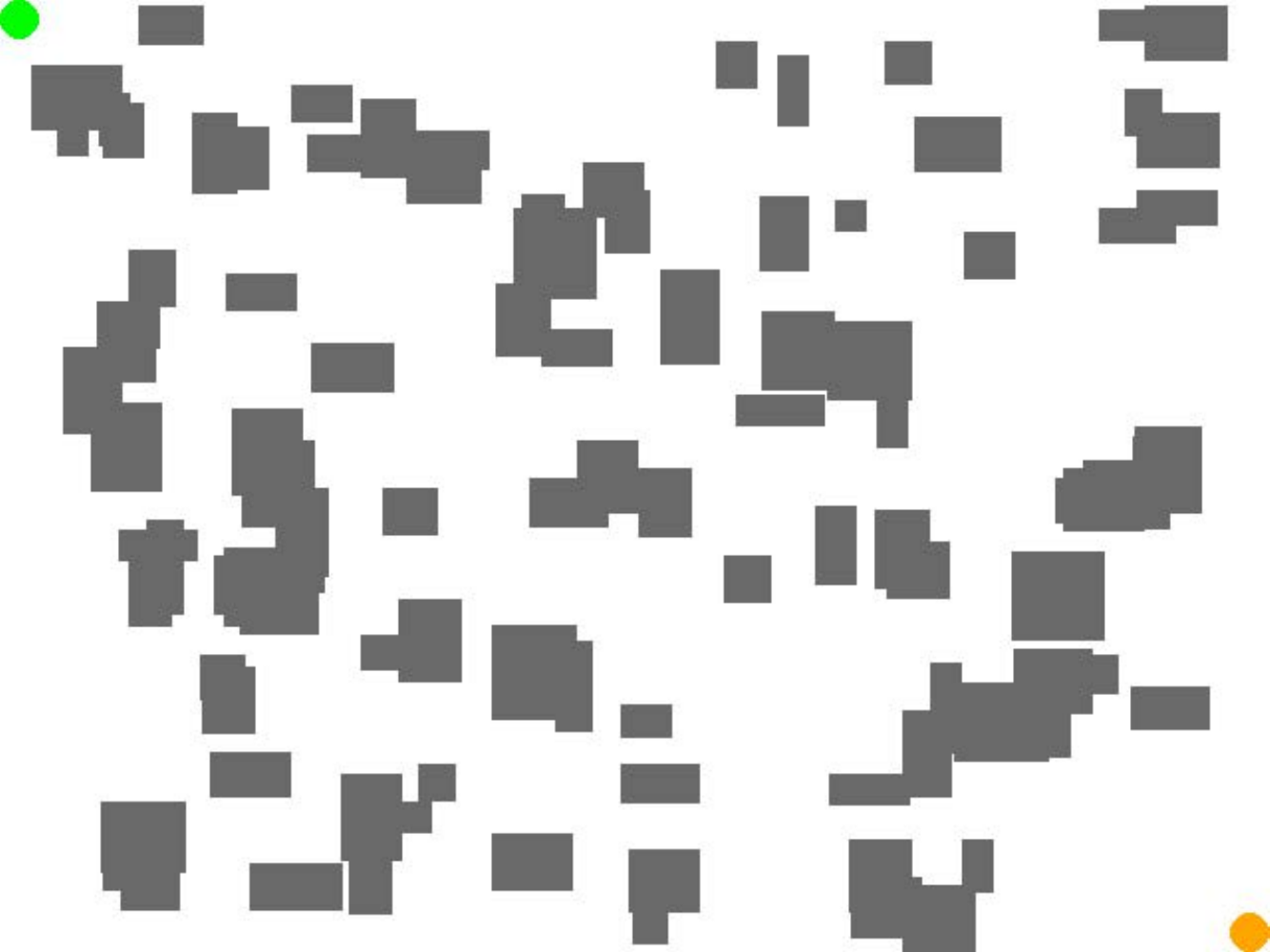}
\includegraphics[width=0.3\textwidth,cfbox=gray 1pt 1pt]{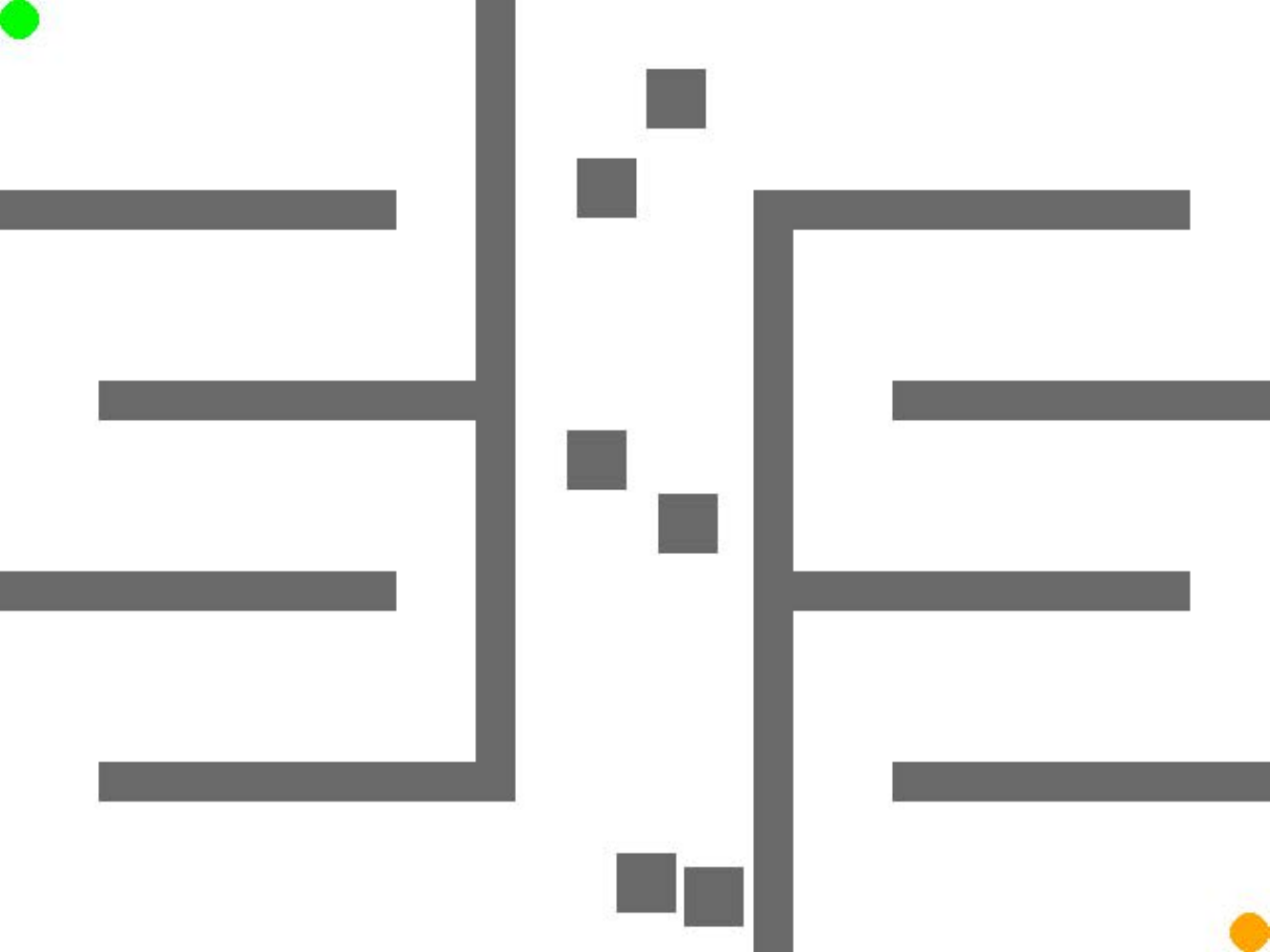}

\vspace{1mm}
\hspace{0.05mm}
\includegraphics[width=0.3\textwidth,cfbox=gray 1pt 1pt]{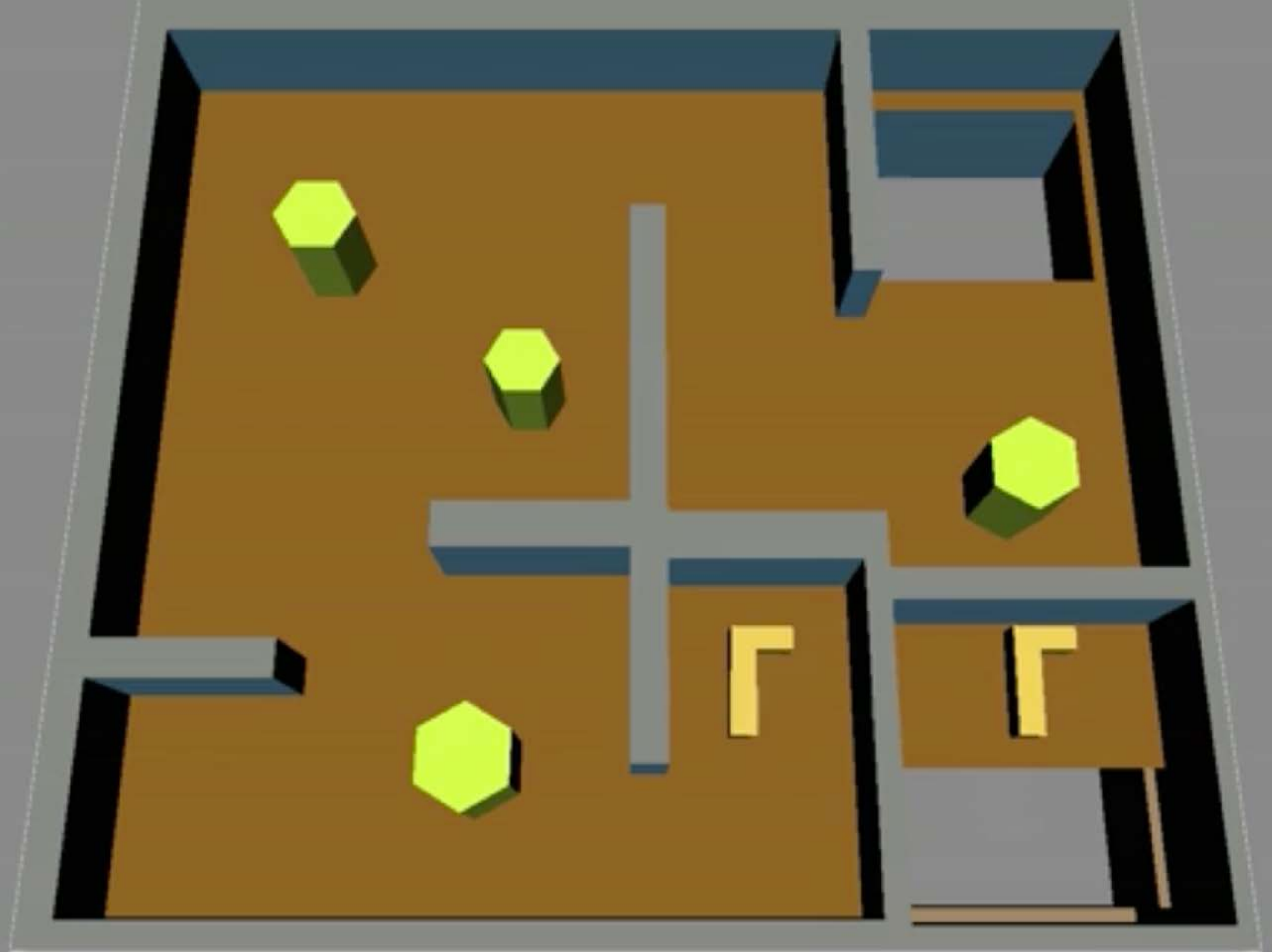}
\includegraphics[width=0.3\textwidth,cfbox=gray 1pt 1pt]{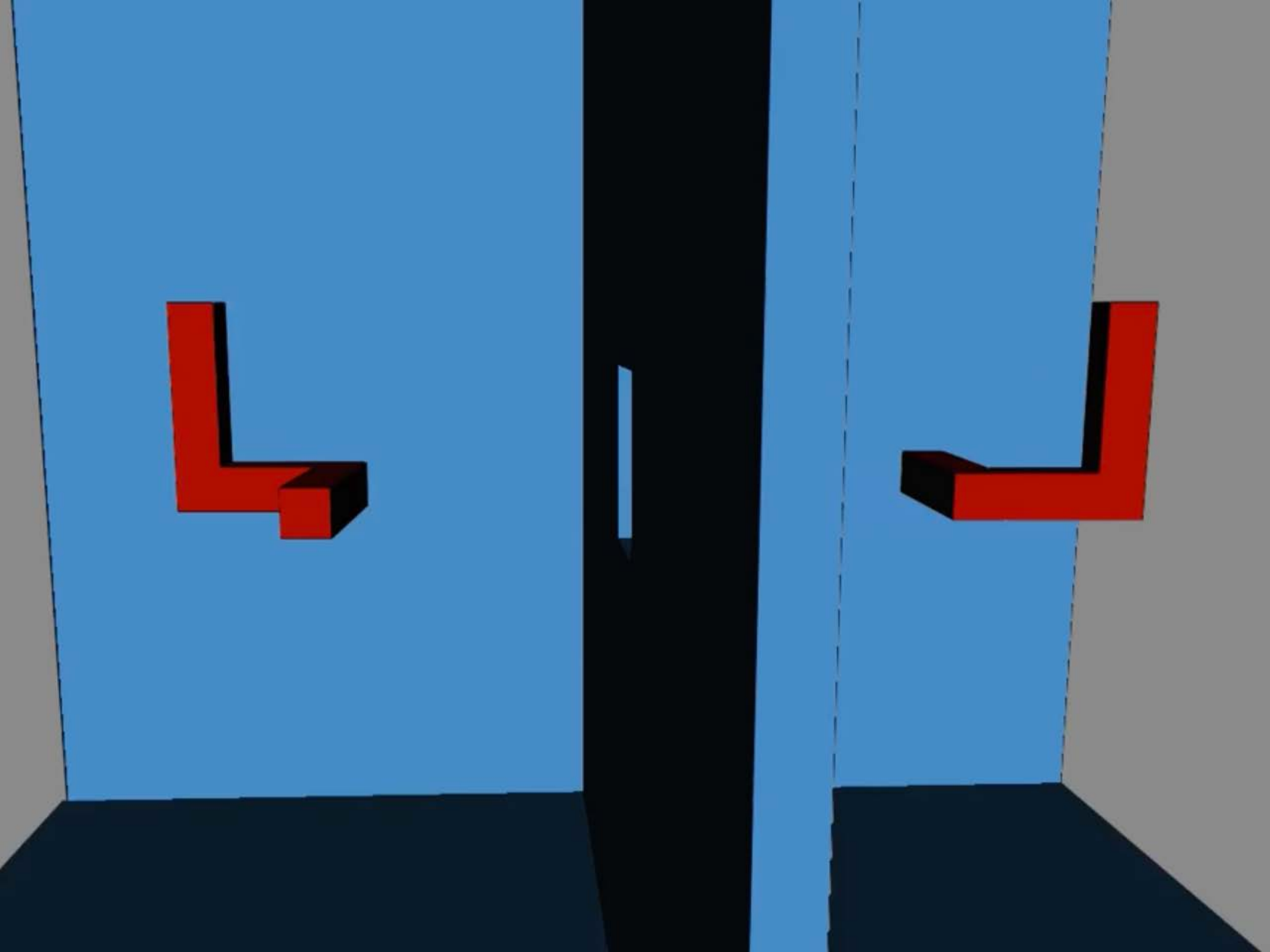}
\includegraphics[width=0.3\textwidth,cfbox=gray 1pt 1pt]{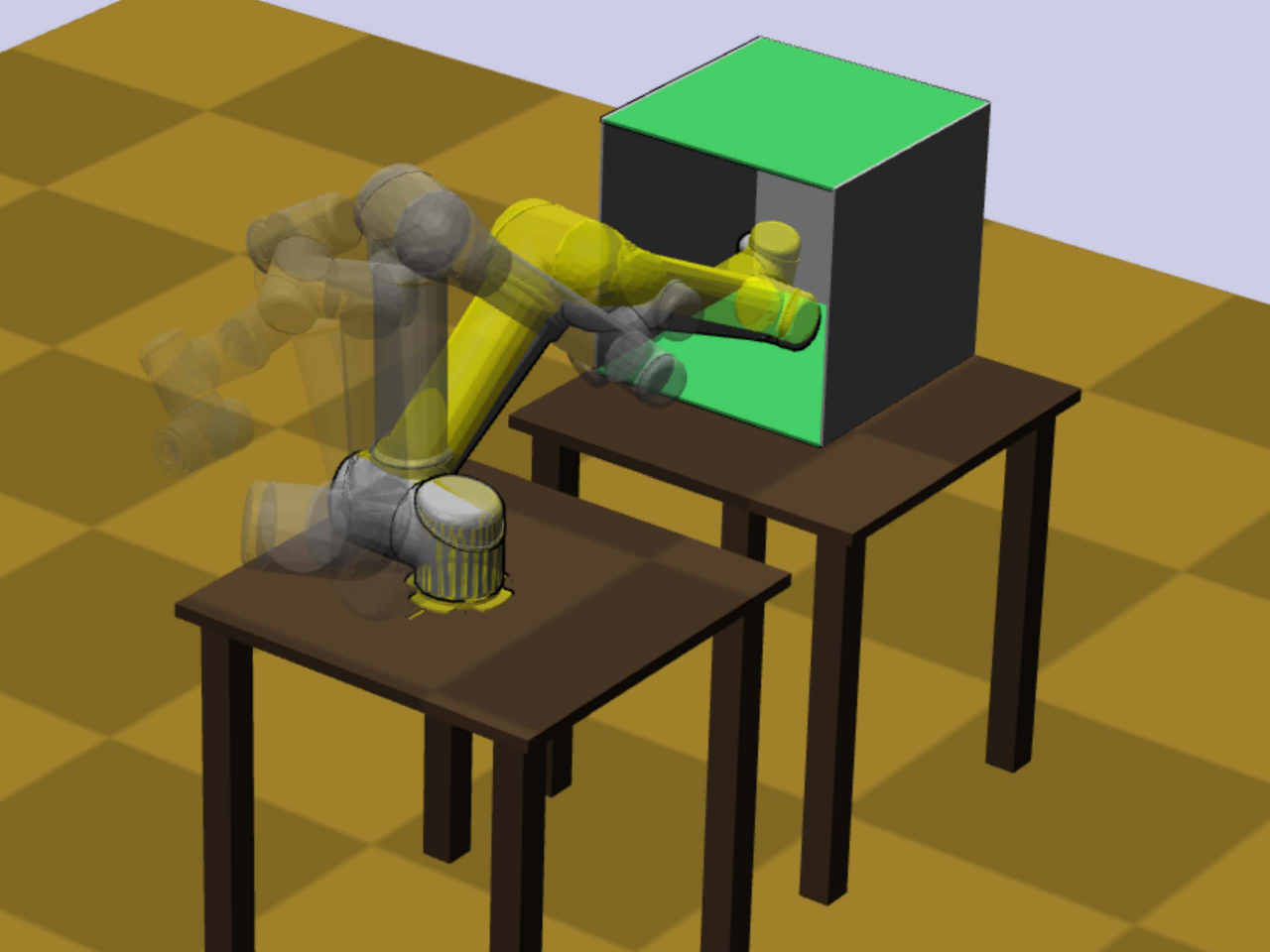}
\caption{Experiments of increasing task complexity: 2D mazes (top) with start (green) and goal (orange) for basic navigation, followed by SE(3) (bottom left/center) for complex object rotation/translation, and 5-D UR5 (bottom right) for constrained robotic manipulation.}
\label{fig:experiments}
\vspace{-10pt}
\end{figure}

\paragraph{2-D maze.}
The goal is to navigate a series of three 2-D mazes depicted in Figure \ref{fig:experiments}. These are designed only to be increasingly difficult to solve for PRM with Uniform sampling before comparing to the other sampling techniques to avoid introducing any bias favoring a specific scheme. The maps are of size 640 by 480 and the agent is assumed to be a disk of radius 6 for collision checking. A single start and end goal is considered and 50 planning attempts are run by level, number of sampled points and sampling technique. The shortest path solver used is $\texttt{A}^*$~\citep{Hart1968}.

\paragraph{OMPL benchmarks}
All of the Uniform, Halton and MPMC sampling schemes are put to the test on experiments from the popular Open Motion Planning Library (OMPL)~\citep{ompl}. The base code PRM implementation is only modified to account for our fixed number pre-sampling. We ensure sufficient compute time to reach failure and run 50 iterations per sampling method, per number of points sampled and per scenario described below:

\textit{1. SE(3) rigid body puzzles. }
The Special Euclidean Group in 3 Dimensions SE(3) is composed of a 3-D translational component and a 3-D rotational component. The latter is often conveniently sampled in quaternion space, with uniform sampling on the 4D unit sphere manifold. In keeping with standard practice, we evaluate samplers' efficient coverage only on the translational 3-D Euclidean space component (discussion on this point provided in Appendix~\ref{app:noneuc}. Two puzzles, Cubicle and Twistycool, are benchmarked from the OMPL. For the latter, we reduce the bounds to allow this harder problem to be solved with the same $N$ values. Both examples can be visualized in Figure \ref{fig:experiments}

\textit{2. $d$-D hypercube corridor. }
On the \( d \)-dimensional hypercube, the valid region is defined such that there exists an index \( k \) with the following constraints: for all dimensions \( i < k \), the $i$-th coordinate must be less than or equal to a threshold edge width \( \lambda \), and for all dimensions \( i > k \), it must be greater than or equal to \( 1 - \lambda \). This results in a valid subspace that resembles narrow passageways along the hypercube edges, leading from one corner of the hypercube to the diagonally opposite corner. For dimension \(d \in \{2,3,10\}\), the value of the edge width is tuned to ensure a meaningful comparison between samplers (respectively $\lambda \in \{0.1, 0.2, 0.37\}$).

\textit{3. $10$-D kinematic chain. }
The experiment features a robot with multiple links in a 10-dimensional configuration space. The robot must navigate through an environment with obstacles represented as line segments. Initially, the robot’s first link is at zero radians, and subsequent links are arranged with a specific angular offset. The goal is to reach a target configuration where the first link aligns nearly with a desired orientation, while avoiding collisions with the environment and itself.

\paragraph{UR5 robot arm}
The UR5 robot arm is a popular collaborative robot with a 6 degree of freedom workspace. In this benchmark, our goal is to reach into a sideways box without contacting the table or the box. We forgo an end effector and thus we do not utilize the last DOF of the robot (which rotates the end effector and in this setting does not effect the workspace of the manipulator). Simulation, sampling, and planning is implemented in the Klamp't software package. We compare our sampling method to sampling from the uniform distribution. For a visualization of the task in simulation see Fig. \ref{fig:experiments}. We also demonstrate our results on real hardware. 

\begin{figure}[t]
\centering
\includegraphics[width=\textwidth]{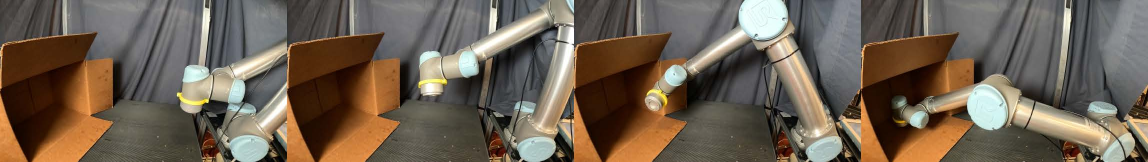}
\caption{Demonstration of a motion plan on hardware. The goal is to reach into the box without making contact.}
\label{fig:demo}
\vspace{-10pt}
\end{figure}

\section{Results}

\subsection{2-D maze}

\begin{wrapfigure}{r}{0.5\textwidth}
    \vspace{-15pt}
    \begin{minipage}{0.5\textwidth}
    \centering
    \includegraphics[width=\textwidth]{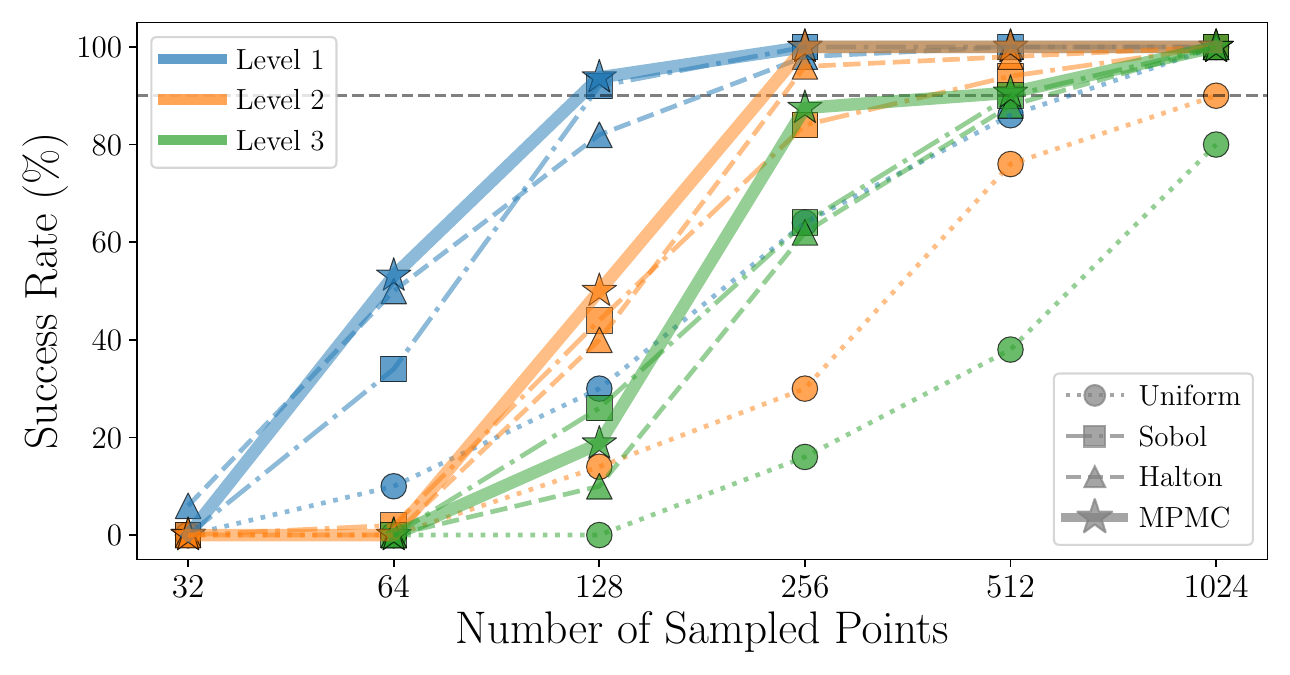}
    \vspace{-15pt}
    \caption{Success rates on the 3 levels of 2-D mazes versus the number of points sampled. Per-level results are grouped by color and the data by sampler is identified by marker and line styles.}
    \vspace{-15pt}
    \label{fig:maze-res}
    \end{minipage}
\end{wrapfigure}

The performance of the different sampling techniques for all three levels of the 2-D mazes are presented in Figure \ref{fig:maze-res}. On all three levels and for all fixed number of points, MPMC offers the best value per sampled point set size on all but two instances (outperformed by Halton by 2\% on level 1 with $N=32$ and by Sobol by 8\% on level 3 with $N=128$). Indeed, MPMC's efficiency superiority is all the more established on the harder level, where it can reach close to 90\% with only 256 points sampled, a feat that takes Halton and Sobol twice the number of points to reach, and a performance level Uniform sampling fails to reach even with 4 times that number.

In fact, Uniform sampling is substantially less efficient as it requires sampling at least 4 times the number of points of low-discrepancy methods to reach the 90\% mark on all levels.
Elsewhere, Sobol and Halton sequences seem to offer comparable performance overall, with relative ranking seemingly contingent on the level and number of points sampled. They often fail to meet the performance of MPMC with significant margins (e.g. 34\% success for Halton on level 1 with $N=64$ against MPMC's 54\%, under 64\% for both Halton and Sobol on level 3 with $N=256$ to MPMC's 88\%).

\subsection{OMPL and UR5 benchmarks}

The benchmarking success rates are provided in Table \ref{tab:results}, along with the average number of valid points sampled by each scheme and its standard deviation.

On the SE(3) Cubicle and Twistycool experiments, MPMC globally achieves the best performance, although with a marginal advantage. This tightness of margins can be linked to the importance of the SO(3) angle quaternion sampling which is done uniformly in all setups. Halton and Uniform schemes appear to be relatively more contingent on the scenario and are harder to clearly rank.

However, MPMC outperforms both Uniform and Halton samplers more consistently and significantly when used on all the samples' components. Indeed, over the 2 and 3 dimensional hypercube examples, MPMC sampling solves around twice as many 
runs as its best competitor with 128 and 256 point regimes and maintains an advantage in the region of 50\% with larger point sets.

The previous observation remains, to a solid extent, valid in higher dimensions. Although relative performance gains are in multiple cases well reduced, MPMC sampling maintains top-performing status. Furthermore, there remains instances where it offers large performance gains.

This is evident on the 10-D kinematic chain test, where MPMC reaches 24\% success rate with 256 samples, compared to only 12\% for Uniform and 14\% for Halton. Similarly, a significant gap is established between MPMC (80\%) and its counterparts (Halton 56\% and Uniform 62\%) with 512 points on the 10-D hypercube benchmark. In the realistic scenario of the UR5 planning task (5-D), MPMC  consistently surpasses Uniform and Halton sampling (3 to 8-fold higher success rates). We also demonstrate that our approach is readily deployed to real hardware (Figure \ref{fig:demo}).

\begin{table*}[t]
\centering
\caption{Benchmark results on the OMPL suite and UR5 robot arm comparing performance in terms of success rate (SR in \%) versus $N$ the number of points sampled \(\{128, 256, 512, 1024\}\). Also provided are the mean and standard deviation of the number of valid milestones in each experiment (denoted ${\lvert V \rvert})$}
\label{tab:results}
\resizebox{\textwidth}{!}{
\begin{tabular}{lclcc|cc|cc|cc}
\toprule
\multirow{2}{*}{\textbf{Experiment}} & \multirow{2}{*}{\textbf{Space}} & \multirow{2}{*}{\textbf{Sampler}} & \multicolumn{2}{c}{\textbf{128}} & \multicolumn{2}{c}{\textbf{256}} & \multicolumn{2}{c}{\textbf{512}} & \multicolumn{2}{c}{\textbf{1024}} \\ \cmidrule(lr){4-11}
 & & & \textbf{SR (\%)} & \textbf{$\mathbf{\lvert V \rvert}$} & \textbf{SR (\%)} & \textbf{$\mathbf{\lvert V \rvert}$} & \textbf{SR (\%)} & \textbf{$\mathbf{\lvert V \rvert}$} & \textbf{SR (\%)} & \textbf{$\mathbf{\lvert V \rvert}$} \\
\midrule
\multirow{3}{*}{Cubicle}
 & \multirow{3}{*}{SE(3)} & Uniform & $0$ & $56.4 \pm 5.0$ & $0$ & $112.9 \pm 7.2$ & $2$ & $221.8 \pm 13.6$ & $26$ & $412.6 \pm 60.8$ \\
 & & Halton & $0$ & $57.4 \pm 3.4$ & $0$ & $112.3 \pm 4.8$ & $4$ & $221.1 \pm 8.9$ & $36$ & $382.4 \pm 98.9$ \\
 & & MPMC & $0$ & $58.8 \pm 3.3$ & $0$ & $112.3 \pm 3.6$ & $\mathbf{10}$ & $221.2 \pm 14.5$ & $\mathbf{46}$ & $367.4 \pm 100.5$ \\
\midrule
\multirow{3}{*}{Twistycool}
 & \multirow{3}{*}{SE(3)} & Uniform & $8$ & $89.2 \pm 14.7$ & $4$ & $180.4 \pm 22.3$ & $26$ & $332.7 \pm 71.6$ & $70$ & $485.0 \pm 221.7$ \\
 & & Halton & $6$ & $89.9 \pm 10.8$ & $\mathbf{16}$ & $171.4 \pm 30.2$ & $20$ & $336.0 \pm 63.0$ & $56$ & $584.1 \pm 193.4$ \\
 & & MPMC & $\mathbf{10}$ & $89.5 \pm 11.5$ & $8$ & $175.2 \pm 27.6$ & $\mathbf{30}$ & $327.7 \pm 67.1$ & $\mathbf{72}$ & $499.3 \pm 212.7$ \\
\midrule
\midrule
\multirow{3}{*}{Hypercube}
 & \multirow{3}{*}{$\mathbb{R}^2$} & Uniform & $24$ & $15.8 \pm 13.5$ & $24$ & $19.1 \pm 24.1$ & $44$ & $56.4 \pm 50.0$ & $60$ & $104.1 \pm 85.6$ \\
 & & Halton & $24$ & $7.9 \pm 11.1$ & $36$ & $19.2 \pm 23.5$ & $36$ & $47.0 \pm 48.9$ & $36$ & $67.5 \pm 90.7$ \\
 & & MPMC & $\mathbf{48}$ & $15.4 \pm 12.4$ & $\mathbf{60}$ & $31.6 \pm 24.4$ & $\mathbf{68}$ & $69.3 \pm 45.5$ & $\mathbf{84}$ & $161.0 \pm 71.0$ \\
\midrule
\multirow{3}{*}{Hypercube}
 & \multirow{3}{*}{$\mathbb{R}^3$} & Uniform & $24$ & $7.9 \pm 7.5$ & $34$ & $17.1 \pm 14.9$ & $44$ & $29.5 \pm 27.4$ & $54$ & $64.1 \pm 51.5$ \\
 & & Halton & $20$ & $5.3 \pm 6.5$ & $32$ & $15.0 \pm 14.2$ & $34$ & $27.6 \pm 27.3$ & $28$ & $42.4 \pm 52.3$ \\
 & & MPMC & $\mathbf{56}$ & $10.3 \pm 6.3$ & $\mathbf{64}$ & $22.5 \pm 12.7$ & $\mathbf{76}$ & $40.6 \pm 22.6$ & $\mathbf{96}$ & $103.1 \pm 20.1$ \\
 \midrule
\multirow{3}{*}{Hypercube}
 & \multirow{3}{*}{$\mathbb{R}^{10}$} & Uniform & $\mathbf{2}$ & $7.0 \pm 2.0$ & $12$ & $9.7 \pm 2.5$ & $62$ & $17.7 \pm 4.3$ & $88$ & $33.9 \pm 5.5$ \\
 & & Halton & $\mathbf{2}$ & $6.6 \pm 1.8$ & $12$ & $10.4 \pm 2.3$ & $56$ & $17.7 \pm 2.9$ & $98$ & $32.6 \pm 3.4$ \\
 & & MPMC & $0$ & $6.3 \pm 1.7$ & $\mathbf{16}$ & $10.7 \pm 2.4$ & $\mathbf{80}$ & $18.9 \pm 1.9$ & $\mathbf{100}$ & $31.1 \pm 2.9$ \\
\midrule
\midrule
\multirow{3}{*}{Kinematic Chain}
 & \multirow{3}{*}{$\mathbb{R}^{10}$} & Uniform & $4$ & $30.2 \pm 4.8$ & $12$ & $57.5 \pm 9.6$ & $22$ & $103.5 \pm 26.3$ & $54$ & $172.5 \pm 64.9$ \\
 & & Halton & $2$ & $30.0 \pm 4.7$ & $14$ & $55.5 \pm 9.1$ & $22$ & $103.7 \pm 21.4$ & $\mathbf{62}$ & $156.0 \pm 64.5$ \\
 & & MPMC & $\mathbf{6}$ & $31.5 \pm 4.3$ & $\mathbf{24}$ & $51.8 \pm 7.5$ & $\mathbf{26}$ & $102.3 \pm 21.5$ & $60$ & $161.7 \pm 71.9$ \\
 \midrule
 \midrule
 \multirow{3}{*}{UR5}
 & \multirow{3}{*}{$\mathbb{R}^{5}$} & Uniform & $0$ & $47.6 \pm 5.8$ & $10$ & $92.2 \pm 7.8$ & $6$ & $184.3 \pm 11.5$ & $12$ & $368.6 \pm 16.1$ \\
 & & Halton & $2$ & $49.4 \pm 6.9$ & $4$ & $93.1 \pm 7.5$ & $14$ & $185.8 \pm 10.8$ & $18$ & $369.2 \pm 15.6$ \\
 & & MPMC & $\mathbf{6}$ & $49.1 \pm 3.0$ & $\mathbf{31}$ & $97.4 \pm 3.5$ & $\mathbf{75}$ & $181.4 \pm 3.8$ & $\mathbf{81}$ & $360.6 \pm 4.3$ \\

\bottomrule
\vspace{-20pt}
\end{tabular}}
\end{table*}

\section{Discussion}

This work introduces the application of MPMC neural network point set generation in motion planning, offering a novel, unbiased approach to sampling. Our proposed method includes a custom training objective specifically designed for high-dimensional configuration spaces, ensuring that the generated points are optimized for complex planning tasks. Moreover, we rigorously show that MPMC point sets offer a tighter upper bound on the distance from the optimal path. 

Through extensive experiments on PRM benchmarks, we demonstrate that our method significantly improves planning efficiency over commonly used sampling methods, especially in high-dimensional and challenging environments. Furthermore, we validate the real-world applicability of our technique by successfully implementing it on a UR5 robot arm, highlighting its potential for deployment in practical robotic systems.


\clearpage

\section*{Limitations and future work}

One limitation of our approach is the need to retrain the MPMC model for each specific number of points $N$ and dimensionality $d$ to ensure optimal performance. This retraining requirement can be computationally intensive, particularly for applications where the configuration space frequently changes. Future work will focus on refining the neural network architecture and training procedure to achieve generalization across both $N$ and $d$, reducing the need for retraining and making the method more flexible across different planning scenarios. Possible avenues to mitigate this limitation are discussed in Appendix~\ref{app:regen}.

Moreover, while we have demonstrated the benefits of MPMC within the PRM framework, future research will explore the potential advantages of integrating this technique into more sophisticated sampling-based planners, potentially extending its impact across a broader class of planning algorithms. Finally, an exciting avenue for future exploration lies in adapting our technique to compute-critical applications, such as acrobatic flight or high-speed driving, where planners require real-time performance and could greatly benefit from efficient, high-quality sampling methods.

\acknowledgments{This work was supported in part by the Postdoc.Mobility grant P500PT-217915 from the Swiss National Science Foundation, the National Science
Foundation EFRI program under grant number 1830901, the Gwangju
Institute of Science and Technology, and the Department of the Air Force
Artificial Intelligence Accelerator and was accomplished under Cooperative
Agreement Number FA8750-19-2-1000. The views and conclusions contained in this document are those of the authors and should not be interpreted
as representing the official policies, either expressed or implied, of the
Department of the Air Force or the U.S. Government. The U.S. Government
is authorized to reproduce and distribute reprints for Government purposes
notwithstanding any copyright notation herein.}


\bibliography{example}  

\clearpage

\begin{appendix}

\section{Appendix}

\subsection{2D point set visualizations}\label{app:2dpts}

\begin{figure}[h]
\centering
\includegraphics[width=\textwidth]{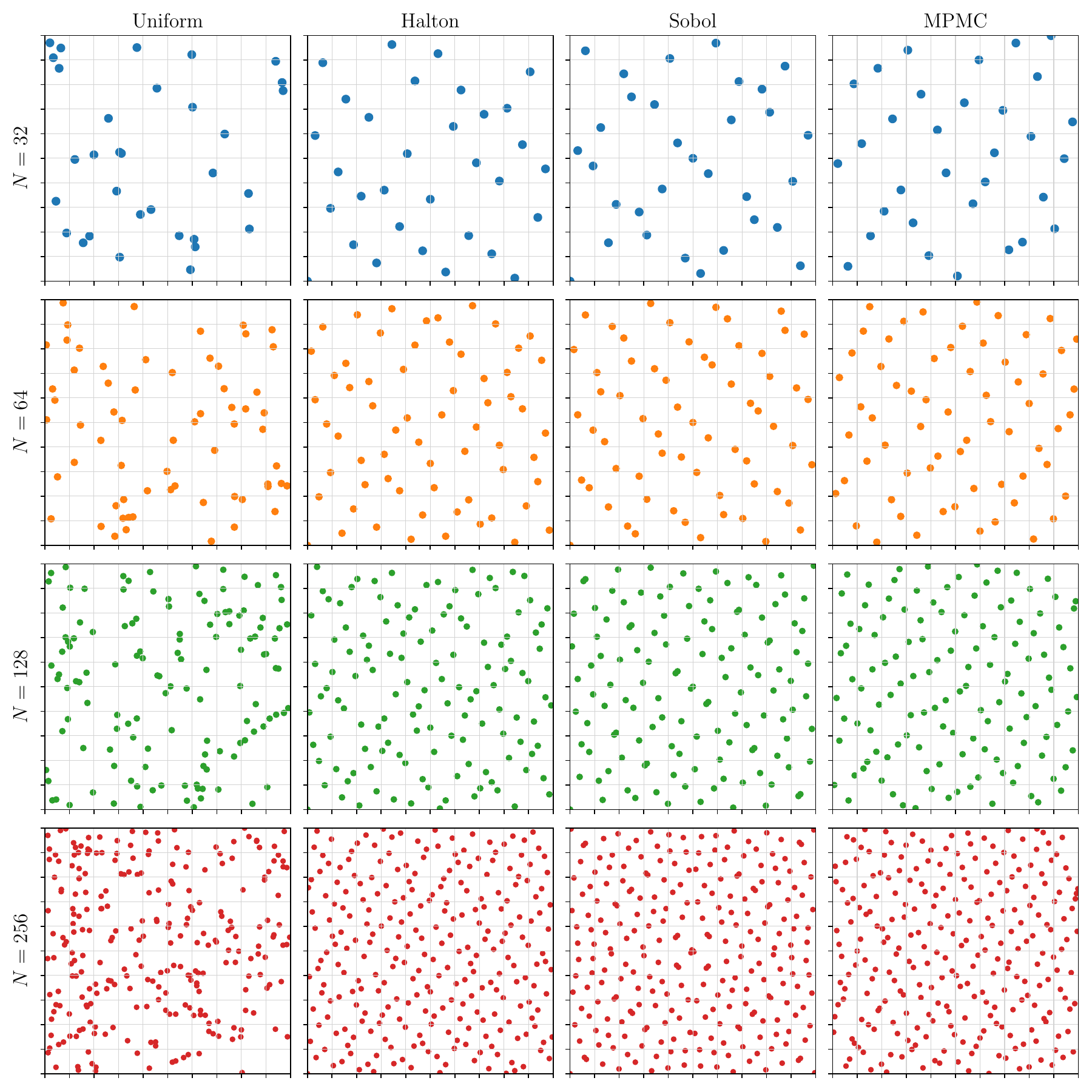}
\caption{Points sampled on the unit square for $N \in \{32,64,128,256\}$ and sampling distributions among Uniform, Halton, Sobol, and MPMC.}
\label{fig:2dpts}
\vspace{-10pt}
\end{figure}

Figure~\ref{fig:2dpts} depicts point sets generated by the various sampling schemes used in the 2D-maze experiments namely, Uniform, MPMC, Halton, and Sobol for different numbers of points. Uniformly sampled points exhibit large gap regions for all numbers of points. The sequential constructions of Halton and Sobol are more evenly spread but are surpassed in discrepancy terms by the learned MPMC points.

\clearpage
\subsection{MPMC point (re)generation}\label{app:regen}

The solution as presented in this work requires retraining a GNN point cloud for each $(N,d)$ pair. Indeed, we focus on optimized point sets (not sequences) to maintain the clearest link between low-discrepancy sampling and improved PRM planning efficiency. There are, however, many practical solutions for mitigating this practicality obstacle and adapting this approach to sequential point generation:
\begin{enumerate}
    \item For sequences over the number of points $N$, we propose training a large MPMC point set that can be transformed into a low-discrepancy sequence using a greedy approach, i.e., successively adding points that minimize the discrepancy over all choices in the point set. This further obviates the need for pruning.
    \item Inductive graph learning is another solution to $N$ generalization, where a GNN can be trained to perform across a varying range of number of nodes (i.e., points). This approach has proven its efficacy in many other contexts.
    \item To deal with varying $d$, a high dimensional point cloud can be used for lower dimensional problems via canonical projection of the MPMC points. Indeed, it has been shown in \citep{mpmc} that MPMC points yield low-discrepancy in particular for lower dimensional projections of the point sets.
\end{enumerate}

\subsection{Dispersion and grids}\label{app:disdis}

Sukharev grids are structured sampling patterns that achieve uniform coverage by placing one point at the center of each cell in an evenly spaced Cartesian grid. In \(d\) dimensions, the domain \([0,1]^d\) is divided into \(k\) equal intervals along each axis, producing \(k^d\) hypercubes, and one sample is placed at the center of each hypercube. Sukharev grids are \(\mathcal{D}_\infty\)-dispersion optimal when \(N = k^d\), with \(k\) an integer. For arbitrary \(N\), however, they can be wasteful: for example, with \(N = 100\) and \(d = 3\), the best the grid can do is \(4 \times 4 \times 4 = 64\) points on the lattice, leaving \(36\) points that do not contribute to an improvement in dispersion. This inefficiency becomes increasingly severe in higher dimensions.

\subsection{The case of non‑Euclidean manifolds}\label{app:noneuc}

MPMC can be extended to general manifolds by developing appropriate objective functions. As an example, MPMC can be extended to $S^n$ by minimizing the total euclidean distance between all pairs of points. However, a general framework of MPMC on manifolds will require a separate investigation to measure the best approach and performance gains in such cases.

\end{appendix}

\end{document}